\documentclass[letterpaper, 10 pt, conference]{ieeeconf}  

\IEEEoverridecommandlockouts                              

\usepackage[font=scriptsize]{caption}  
\usepackage{url}
\usepackage{hyperref}
\usepackage{url}
\usepackage{balance}
\usepackage{amsmath}
\usepackage{amssymb,bm}
\usepackage{amsfonts}
\usepackage{enumerate}
\usepackage{tabulary}
\usepackage{textcomp}
\usepackage{multirow}
\usepackage{cite}
\usepackage{lipsum}  
\usepackage{graphicx}
\usepackage{epstopdf}
\usepackage[misc]{ifsym}
\usepackage{color}
\usepackage{xcolor}
\usepackage{xxcolor}
\usepackage{pgf}
\usepackage{pgfplots}
\usepackage{tikz}
\usepackage{float}
\usepackage{lipsum}
\usepackage{tabularx}
\usepackage{lipsum}

\usepackage{booktabs} 
\usepackage{array}    
\usepackage{graphicx} 
\newcolumntype{P}[1]{>{\centering\arraybackslash}p{#1}}
\pgfplotsset{compat=1.18}

\usepackage{xparse}

\NewDocumentCommand\bbm{}{ \begin{bmatrix} }
\NewDocumentCommand\ebm{}{ \end{bmatrix} }

\NewDocumentCommand\LieGroupSE{m}{ \mathrm{SE}(#1) }

\title{\LARGE \bf
UAV See, UGV Do: Aerial Imagery and Virtual Teach Enabling Zero-Shot Ground Vehicle Repeat 
}

\author{Desiree~Fisker$^{1}$, Alexander~Krawciw$^{1}$, Sven~Lilge$^{1}$, Melissa~Greeff$^{2}$, and Timothy~D.~Barfoot$^{1}$
\thanks{*This work was supported by the University of Toronto Robotics Institute and the Defense Research Development Canada}
\thanks{$^{1}$Desiree Fisker, Alexander Krawciw, Sven Lilge, and Timothy D. Barfoot are with the University of Toronto Institute of Aerospace Studies,
        4925 Dufferin St, North York, ON M3H 5T6, Canada
        {\tt\small \{desiree.fisker, alec.krawciw, sven.lilge\}@mail.utoronto.ca, tim.barfoot@utoronto.ca}}%
\thanks{$^{2}$Melissa Greeff is with Ingenuity Labs and the Department of Electrical and Computer Engineering, Queen's University,
        99 University Ave, Kingston, ON K7L 3N6, Canada
        {\tt\small melissa.greeff@queensu.ca}}%
}

\begin{document}

\maketitle
\thispagestyle{empty}
\pagestyle{empty}

\begin{abstract}
This paper presents Virtual Teach and Repeat (VirT\&R): an extension of the Teach and Repeat (T\&R) framework that enables GPS-denied, zero-shot autonomous ground vehicle navigation in untraversed environments. VirT\&R leverages aerial imagery captured for a target environment to train a Neural Radiance Field (NeRF) model so that dense point clouds and photo-textured meshes can be extracted. The NeRF mesh is used to create a high-fidelity simulation of the environment for piloting an unmanned ground vehicle (UGV) to virtually define a desired path. The mission can then be executed in the actual target environment by using NeRF-generated point cloud submaps associated along the path and an existing LiDAR Teach and Repeat (LT\&R) framework. We benchmark the repeatability of VirT\&R on over 12 km of autonomous driving data using physical markings that allow a sim-to-real lateral path-tracking error to be obtained and compared with LT\&R. VirT\&R achieved measured root mean squared errors (RMSE) of 19.5 cm and 18.4 cm in two different environments, which are slightly less than one tire width (24 cm) on the robot used for testing, and respective maximum errors were 39.4 cm and 47.6 cm. This was done using only the NeRF-derived teach map, demonstrating that VirT\&R has similar closed-loop path-tracking performance to LT\&R but does not require a human to manually teach the path to the UGV in the actual environment. 
\end{abstract}


\section{INTRODUCTION}
Enabling a higher level of autonomous navigation in remote, harsh, and potentially hazardous environments is a critical objective for many Unmanned Ground Vehicle (UGV) operations, as minimizing human presence in such scenarios reduces risk and lowers costs. Visual Teach and Repeat (VT\&R) \cite{furgale_visual_2010}, is a complete autonomy stack that enables long-range navigation along previously taught routes, demonstrated on a UGV with 3D-LiDAR \cite{sehn_along_2022, burnett_are_2022, krusi_driving_2017}, Radar \cite{qiao_radar_2024}, and RGB vision sensors \cite{furgale_visual_2010}, as well as on a UAV with an RGB vision sensor \cite{bianchi_uav_2021, patel_visual_2020}. While Teach and Repeat (T\&R) has demonstrated considerable success, it currently requires a human operator to manually guide the vehicle in the environment during the teaching phase to create a map and ensure traversability.

\begin{figure}[t]
    \centering
    \includegraphics[width=1.0\columnwidth]{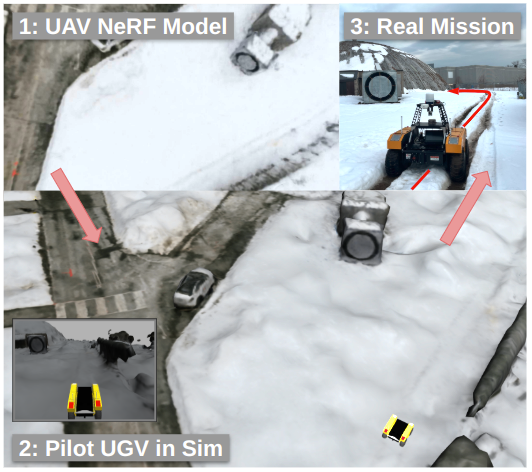}
    \caption{Step 1: UAV NeRF Model, involves flying a drone to capture aerial imagery so a NeRF model can be trained to represent the target environment (pictured is the UTIAS Mars Dome). Then, Step 2: Pilot UGV in Sim, uses a mesh extracted from the NeRF model to virtually pilot a robot through the environment. Lastly, Step 3: Real Mission, sees the UGV autonomously execute the virtually taught mission in the actual environment by matching live LiDAR scans to NeRF-generated point cloud submaps.}
    \label{fig:main_pic}
\end{figure}

Neural Radiance Fields (NeRFs) \cite{mildenhall2021nerf} have emerged as an effective technique for novel viewpoint synthesis. Notably, NeRF models can reconstruct point clouds with accuracy comparable to LiDAR scans \cite{zhang_nerf-LiDAR_2024}, and capture increasingly larger scenes with greater detail and reduced training times \cite{tancik_block-nerf_2022, turki_mega-nerf_2022, tancik_nerfstudio_2023}. Loc-NeRF \cite{maggio_loc-nerf_2022} demonstrates the ability to perform real-time, vision-based global localization using a pre-trained NeRF model. However, our goal is to operate in larger, outdoor environments by using vision and LiDAR sensors on a UAV and UGV respectively, without a global reference. 

In this paper, we extend the Visual Teach and Repeat 3 (VT\&R3) framework \footnote{\url{https://github.com/utiasASRL/vtr3}} \cite{furgale_visual_2010} by integrating multiple platforms \cite{Warthog, DJIPhantom4} into a novel architecture for generating teach maps that bridges a gap between sensor modalities \cite{DJIPhantom4, ouster_128} to exploit their respective benefits. From a NeRF-rendered scene, we extract both a high-quality mesh that is used for virtually piloting a ground vehicle in simulation, as well as a dense point cloud that is used to localize the ground vehicle in the target environment. Our proposed framework allows for different paths to be taught in simulation, creating pose graphs that integrate directly with the LiDAR Teach and Repeat (LT\&R) pipeline, to enable an autonomous repeat of the virtually taught route in the real environment.

The proposed Virtual Teach and Repeat (VirT\&R) system is implemented, tested, and evaluated on a Clearpath Warthog UGV. Autonomous repeats spanning over 12 km were conducted across a variety of environments, where Fig. \ref{fig:main_pic} illustrates the process for creating a virtual teach map and using a Clearpath Warthog to execute a mission without previously entering the environment. Path-tracking performance is then assessed and compared with an existing LT\&R approach used previously \cite{qiao_radar_2024}. To the best of the authors’ knowledge, this work presents the first demonstration of zero-shot closed-loop driving in outdoor environments by combining volumetric rendering of aerial imagery and a LiDAR sensor. 

\subsection{Related Work}
\subsubsection{Teach and Repeat}
The T\&R navigation algorithm includes a teaching phase where local submaps are constructed through manual driving and followed in an autonomous repeat phase.
The pose graph framework of T\&R uses any sensor capable of providing a geometric world representation for localization \cite{furgale_visual_2010, sehn_along_2022, burnett_are_2022, krusi_driving_2017, chen_self-supervised_2022, gridseth_keeping_2022}, and in this paper, we exploit this capacity to make use of the NeRF-generated point cloud.
By constructing a topometric pose graph, T\&R retains the ability to achieve metric localization to the path without global consistency, which has been demonstrated on a UAV with vision sensors as well \cite{patel_visual_2020, bianchi_uav_2021}. However, the only attempts made towards cross-modal T\&R have been with Radar and LiDAR sensors \cite{lisus2024pointingwayrefiningradarlidar, burnett_are_2022}, and these still required a manual teach phase in the target environment. 

\subsubsection{Neural Radiance Fields}
NeRFs represent scenes as continuous volumetric functions that can be queried for color and density at viewpoints not included in training data. FlyNeRF \cite{dronova2024flynerf} is an example of drone-based data acquisition for high-quality 3D scene reconstruction. Large-scale NeRF representations have been addressed by Block-NeRF \cite{tancik_block-nerf_2022} and Mega-NeRF \cite{turki_mega-nerf_2022}, which scale NeRF models for city-sized environments by decomposing them into multiple sub-networks, improving rendering efficiency and fidelity. 

NeRF’s potential for localization and SLAM has also been widely explored. NeRF-SLAM \cite{rosinol_nerf-slam_2022} integrates NeRF into a monocular SLAM framework, enhancing real-time scene reconstruction. NICE-SLAM and NICER-SLAM \cite{zhu_nice-slam_2022,zhu_nicer-slam_2023} extend functionality by using monocular, multi-level local geometric cues for improved mapping and tracking. NICE-SLAM was also improved upon by incorporating depth uncertainty and motion information to obtain better tracking accuracy and robustness \cite{lisus_towards_2023}. 

Additionally, several works have focused on real-time localization with NeRFs, using Monte Carlo localization for vision-based global pose estimation \cite{maggio_loc-nerf_2022}, neural signed distance functions to enhance LiDAR odometry and mapping \cite{deng_nerf-loam_2023}, and couplings with visual-inertial odometry pipelines for pose estimation \cite{katragadda_nerf-vins_2024, han_nvins_2024}. Rapid-Mapping \cite{zhang_rapid-mapping_2024} integrates LiDAR and visual data to create high-fidelity scene reconstructions while maintaining real-time performance. Most applicable to our work, LocNDF \cite{wiesmann_locndf_2023} learns a Neural Distance Field to create a localization map from LiDAR data. However, LocNDF does not directly employ NeRFs to produce their localization layer, does not incorporate a different sensor for scene captures, and relies on a rough initial GPS location for the Iterative Closest Point (ICP) algorithm. Another approach related to our work is NeuRAD \cite{tonderski_neurad_2024}, which uses NeRFs to create a simulation tool for testing autonomous driving sensors. While they leverage NeRFs to extract sensor readings in simulation, they do not perform real-world localization. In contrast to related work, we develop a framework that leverages NeRFs both for simulation generation and map generation for real-time localization with a ground vehicle in a physical environment.

\subsubsection{UAV-UGV Cooperation}
Diverse UAV-UGV collaboration strategies have been employed for many aspects of autonomous operations, such as estimating a drone’s position relative to a ground vehicle \cite{hausberg_relative_2020}, and a SLAM framework with Ultra Wide Band (UWB) ranging units to enhance UGV localization accuracy \cite{qian_localization_2024}.

Combined mapping efforts such as \cite{qin_autonomous_2019} involve a UAV-UGV system that employs a two-layered exploration strategy of coarse UGV-based mapping with fine UAV-based 3D reconstruction. Similarly, a cooperative exploration method where a UAV provides aerial situational awareness and relative positioning using fiducial tag tracking has been developed \cite{hood_birds_2017}. UAV-UGV path planning and obstacle mapping have also seen considerable advancements with a multi-UAV stereo vision system for obstacle mapping \cite{kim2014multi}, and a UAV-assisted path planning framework where aerial images are used to detect obstacles and generate safe routes \cite{lakas_framework_2018}. However, none of these cooperative methods use the UAV exclusively for mapping offline prior to UGV operation, with one method using satellite imagery for NeRF modeling beforehand in a similar initial cost-mapping phase \cite{dai_neural_2023}.

Other existing works leverage a UAV to scan terrain ahead of a moving UGV \cite{gilhuly_robotic_2019}, integrate aerial and ground LiDAR for collaborative risk mapping to enable real-time terrain assessment for off-road UGVs \cite{wang_aerial-ground_2023}, and employ a UAV-guided UGV navigation system that optimizes guidance viewpoints and trajectory planning in cluttered environments \cite{yang_guidance_2024}. However, these works focus on simultaneous UAV and UGV operation and communication, not on creating localization maps for later UGV navigation. 

\subsection{Contributions}
We present a complete Virtual T\&R framework and evaluate its path-tracking performance with LiDAR T\&R \cite{burnett_are_2022} as a baseline. The contributions of this work are:
\begin{itemize}
\item a Virtual Teach and Repeat framework that expands current T\&R capabilities for creating taught routes; 
\item an experimental methodology for a quantitative comparison of the accuracy and repeatability of T\&R using NeRF-generated teach maps; 
\item and demonstrated experimental repeatability on over 12 km of autonomously driven routes.
\end{itemize}

The remainder of this paper is organized as follows. Section~\ref{sec:methods} describes the methodology and implementation of the VirT\&R pipeline. The experiments are outlined in Section~\ref{sec:experiments}, and Section~\ref{sec:results} summarizes the results and compares them to LT\&R as a baseline. Finally, Section~\ref{sec:discussion} analyzes our findings and the lessons learned. Section~\ref{sec:conclusion} concludes the paper and suggests future pipeline improvements.

\section{Methodology}
\label{sec:methods}
The online repeating phases of VirT\&R and LT\&R are identical as we present an expansion to the algorithm \cite{qiao_radar_2024} by enabling a new way to teach, and existing downstream modules of T\&R that are run live during a repeat handle motion planning and control. Initial scene capture consists of flying a small UAV at various altitudes to gather different viewpoints of a scene for NeRF training data. A detailed mesh of the environment produced from this NeRF model is used to create a simulator for driving the UGV to obtain the desired path through the environment as relative transformations. Subsequently, a dense point cloud from the NeRF model is used to construct a sequence of small submaps, which become associated with vertices along the desired path. Our method of teaching paths for the UGV enables multiple new paths to be virtually piloted through the captured scene without the UGV entering the environment beforehand. The capability of teaching multiple paths for the UGV to execute, given only one instance of a UAV capturing the scene, is a key benefit of our method. 

\subsection{Virtual Teach Pipeline}
The VirT\&R pipeline, illustrated in Fig. \ref{fig:pipeline}, outlines the key steps involved in enabling a UGV to navigate autonomously in an environment it has never physically traversed. The VirT\&R pipeline consists of three main phases: a data capture and offline pre-processing phase to obtain a scene reconstruction and localization layer, a simulated pilot driving phase to create the desired path, and an online execution phase where the robot completes a real mission. The offline VirT\&R map creation method is as follows:

\subsubsection{Scene Construction}
The offline phase begins with UAV-based data collection, where aerial imagery is captured with a view looking straight down from approximately 20 m, looking forward at approximately 5-10 m, and looking at a 45\textdegree{} angle inwards on the scene at approximately 15 m. These images serve as the input for generating a 3D model of the environment and can optionally be post-processed with associated GPS poses so the NeRF model retains an exact real-world scale. However, this step is not necessary for the pipeline, as recovering a scaled scene can be done with methods such as the ICP algorithm. The reconstruction and optional GPS scaling are done using Colmap \cite{schoenberger2016sfm}, which creates the inputs for the NeRF model that correspond to the training images and took 43 minutes for the largest scene with 1555 960x540 images. The modeled scenes range from approximately 80 x 80 m to 150 x 100 m.

\subsubsection{NeRF Model Training}
We utilize Nerfstudio's \cite{tancik_nerfstudio_2023} seminal model, `Nerfacto', for training the scenes used for navigation, but any volumetric rendering or 3D reconstruction that produces high-quality, dense results could also be used. This model combines Instant-NGP’s \cite{mueller2022instant} hash grid encoding with a small multi-layer perception (MLP), enabling rapid training, large-scale scene modeling, and preservation of the input reconstruction's coordinate system. 

\begin{figure}[t]
\vspace*{1mm}
    \centering
    \includegraphics[width=1.0\columnwidth]{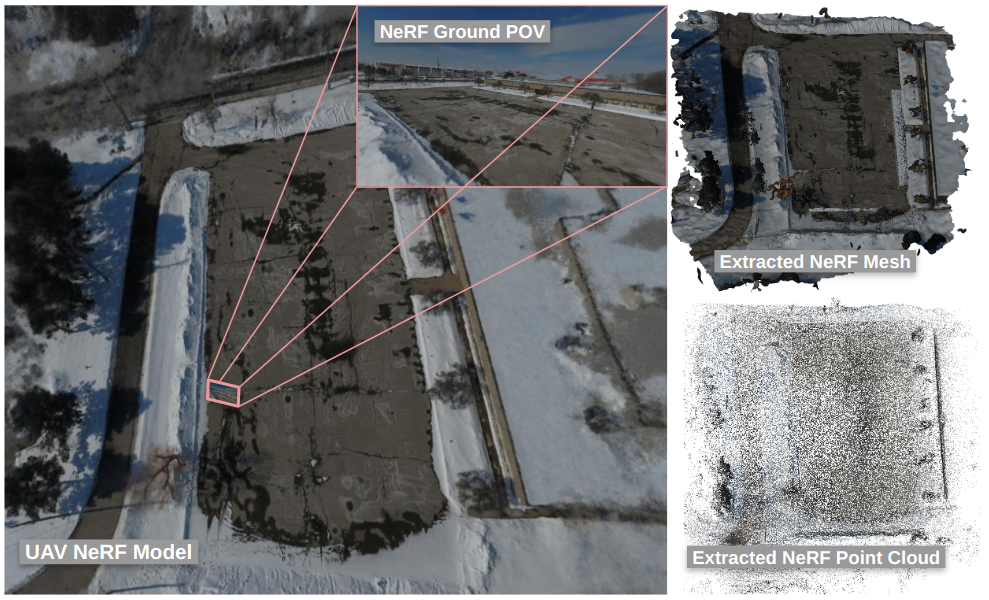} 
    \caption{Visualization of all NeRF-generated components that make up the pipeline for the same scene. Left: Image of the trained Nerfacto model of the UTP Parking Lot Loop with a key frame highlighted and exploded in red to showcase the quality of the novel viewpoint synthesis that allows for dense and geometrically accurate meshes and point clouds to be extracted. Top Right: The corresponding NeRF Mesh. Bottom Right: The corresponding NeRF Point Cloud.}
    \label{fig:combined}
    \vspace{-5mm}
\end{figure}

The training is done offline using UAV imagery on an NVIDIA GeForce RTX 3090 GPU, and took 22 minutes for the largest scene with 1555 960x540 images.
We obtain two key representations from the NeRF model once training has finished: a dense point cloud and a textured mesh that can be seen in Fig. \ref{fig:combined} for the UTP Parking Lot Loop. The resulting representations of the environment are essential for subsequent localization and navigation tasks, as the point cloud provides a rich LiDAR-like representation of the environment, while the mesh facilitates visually accurate simulation-based trajectory planning.

The mesh is imported into Gazebo \cite{noauthor_gazebo_nodate} after being extracted from the NeRF model with the point cloud. The virtual teach pass can then be driven using a joystick controller and saved as the absolute starting position and subsequent relative transformations in the Gazebo world frame. This simulated environment allows an operator to define a trajectory using a physics-based model of the Clearpath Warthog, replicating real-world driving and terrain assessment. 

\subsubsection{NeRF Point Cloud Path Association}
To integrate NeRF-based localization with existing LT\&R frameworks, the extracted point cloud is processed into submaps. These submaps are small cylindrical croppings of the NeRF point cloud, which allow the UGV to localize itself relatively using its onboard LiDAR sensor.
The relative transformations of the path from Gazebo through the mesh are compounded using $\LieGroupSE{3}$ rigid body transformations, and the result is a path from Gazebo traversing the NeRF point cloud, providing the trajectory to link together submaps saved at certain distance and orientation thresholds. The relative transformation between each saved submap is used to decide which submap should be used for localization during a repeat. It is worth noting that we remain true to the topometric map philosophy of T\&R since it is not critical that our NeRF model is globally metrically accurate.

\begin{figure*}[t]
\vspace*{1mm}
    \centering
    \includegraphics[width=1.0\textwidth]{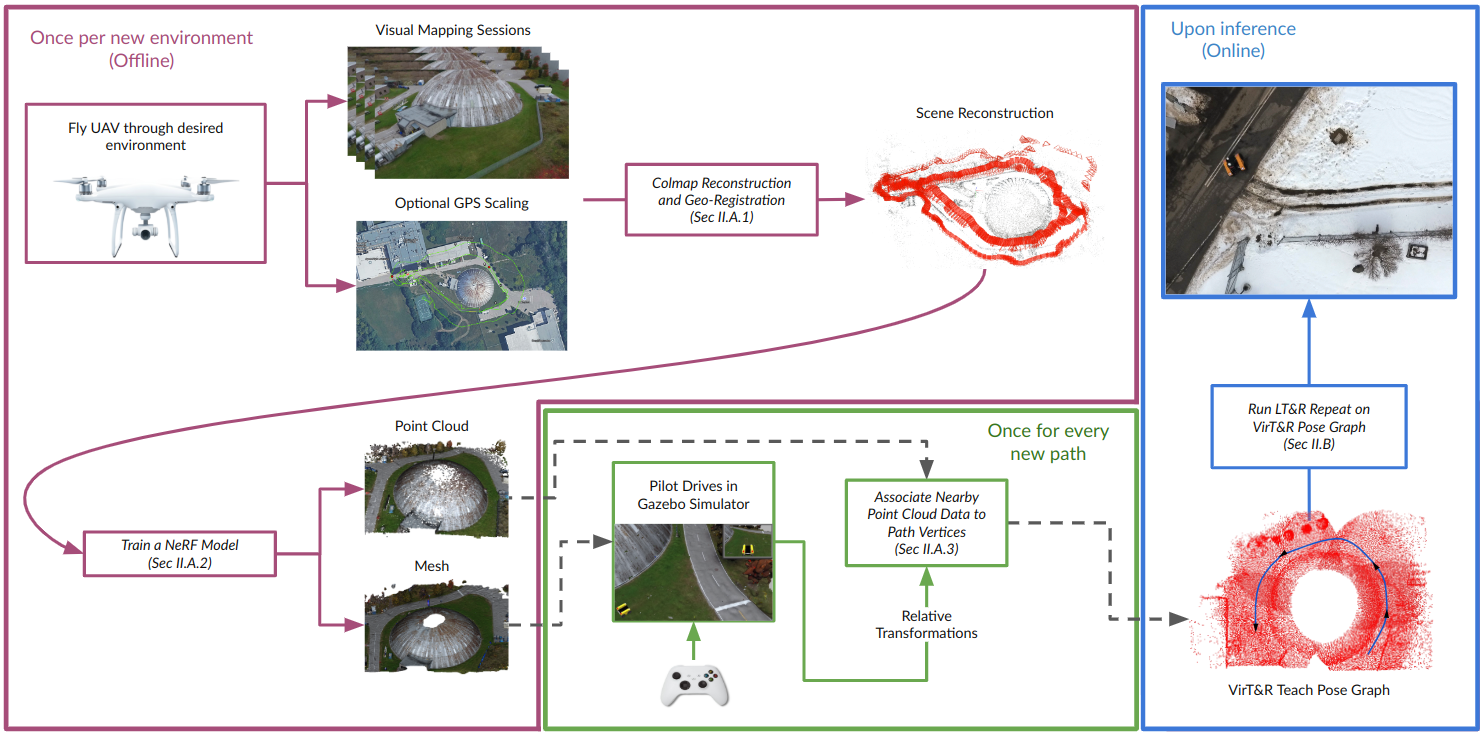} 
    \caption{Overview of the UAV-based mapping and UGV navigation pipeline. Offline Mapping (magenta): A UAV captures images and GPS data, which are processed using Colmap for reconstruction, followed by NeRF training to generate point clouds and meshes. Path Planning (green): The desired path is piloted in simulation, and the point cloud is used to associate nearby points with path vertices. Online Navigation (blue): A UGV follows the path using LT\&R for real-world localization and navigation.}
    \label{fig:pipeline}
    \end{figure*}

\subsection{Autonomous Repeat}
The localization layer in T\&R is the geometric world representation that the UGV uses to localize itself in an environment. VirT\&R derives this layer from the NeRF-generated point cloud instead of LiDAR data, providing a similarly dense and locally metrically accurate 3D structure of the scene. During deployment, the UGV localizes itself by registering its live LiDAR scans to NeRF-derived submaps using the ICP point-to-plane algorithm \cite{Chen1992}. Once localization is established, the UGV follows the virtual teach path, repeating as if it were physically taught with traditional T\&R.

\subsection{Evaluation and Ground Truth}
The method for evaluating accuracy and path-tracking performance in VirT\&R differs from prior work \cite{furgale_visual_2010, chen_self-supervised_2022, gridseth_keeping_2022, sehn_along_2022, burnett_are_2022} because the teach phase is conducted entirely in a virtual environment. As such, we do not have the GPS data for teach paths commonly used to quantify path-tracking error (PTE).

In an attempt to produce `pseudo-GPS' data for evaluation, we recovered the path taught with the Gazebo simulation in an Earth Centered, Earth Fixed (ECEF) coordinate frame using the GPS data from the UAV that gave the accurate scale to the NeRF model. GPS data for repeats to the virtual teach map was also recorded using a NovAtel OEM7 GPS \cite{novatelGPS} on all routes. Analysis of the resulting GPS-measured PTE revealed drastically elevated values that did not align with our qualitative observations or the physical markings. It is likely that multiple sources of error, such as GPS receiver differences, a weeks-long gap between GPS measurements, Colmap and NeRF reconstruction inaccuracies, and the lack of GPS post-processing on UAV data, were compounded in creating the pseudo-GPS data. Thus, it was deemed impractical for use in measuring the actual PTE from control and localization.

\begin{figure}[ht!]
    \centering
    \includegraphics[width=1.0\columnwidth]{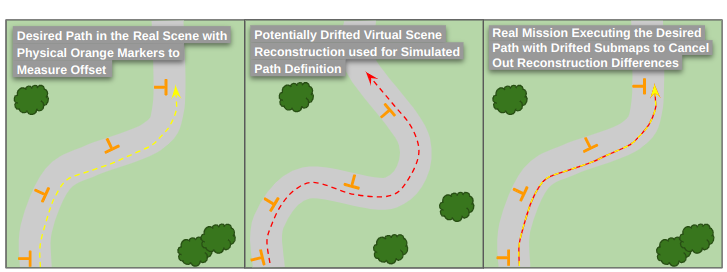} 
    \caption{A visualization of a desired real world path with spray paint markings, an exaggeration of what potential drift could do to the virtually-generated mesh and point cloud (and thus the virtual teach path), and the desired path being repeated due to equally applied drift so the correct teach path is recovered in the real world.}
    \label{fig:drift}
\end{figure}

The lack of GPS data for conventional evaluation led us to establish a physical marking-based sim-to-real path-tracking assessment method. Physical markings in the environment are the most suitable tool to assess the UGV's ability to track the virtual teach path because they ensure the PTE accounts for our own errors in piloting the UGV in simulation, as well as localization and control errors, not any that may result from the scene reconstruction process. This is due to the markings appearing in the drone imagery, the Colmap reconstruction and NeRF model, and the photo-textured mesh, meaning any drift or distortion from scene reconstruction is applied equally to the markings, the virtual teach path driven alongside them, and the point cloud used for localization. A visualization of how reconstruction error is implicitly canceled out can be seen in Fig. \ref{fig:drift}, where the use of local, potentially drifted submaps made from the virtually-generated point cloud are localized to the real environment to enable the desired virtually taught path to be repeated.

\begin{figure}[t]
    \centering
    \includegraphics[width=1.0\columnwidth]{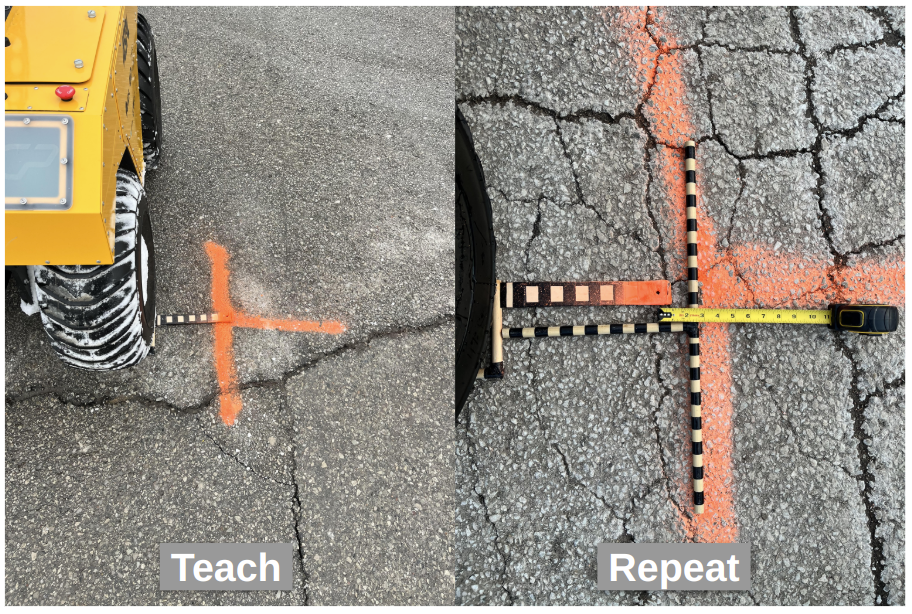}
    \caption{Left: Use of the jig to repeatedly make spray paint marks on the ground next to the rear right tire at some offset for measurement and assessment of path tracking accuracy during an LT\&R teach path. Right: Measurement to the spray paint mark while the UGV is repeating to a virtual teach path. Repeats were also done using LT\&R to obtain measurement offsets to directly compare the two methods.}
    \label{fig:jig}
\end{figure}

To implement this methodology, an LT\&R teach pass was conducted on the Mars Dome and Parking Loops during which the robot was driven and periodically stopped to spray paint marks near the rear-right tire. These marks were then captured in the UAV footage, scene reconstruction, and mesh, so they could be virtually piloted alongside of as closely to the distance they were painted from the tire as possible, to generate a comparable VirT\&R teach pass for offsets to be measured during a real repeat. The jig used to apply and measure the markings in the real world is shown in Fig. \ref{fig:jig}.

To further validate performance, we repeated to the LT\&R and VirT\&R teach maps respectively, so that mark offset measurements could be taken for both algorithms, enabling a direct comparison between the two methods. Although aligning the UGV with the marks in the simulation is inherently imperfect, this challenge mirrors the real-world difficulty of precisely following a physical reference path. However, it does prevent us from comparing the exact same paths, as the virtually driven path is likely not exactly what the LT\&R path used to place the markings is. To account for this human error, a ruler model was attached to the Warthog in Gazebo, allowing measurements from the Warthog’s tire to the spray paint mark to be taken in simulation as well. We can better quantify the PTE by subtracting the error attributable to pilot imprecision from the VirT\&R repeat offset measurements, thereby providing a more accurate benchmark of performance and the sim-to-real gap.

\section{Experiments}
\label{sec:experiments}
Several environments with varying levels of structure were chosen to show that the route-following performance of VirT\&R is consistent. Performance is assessed with the aforementioned methodology to compare VirT\&R with LT\&R, and the deviation between two example repeats of all VirT\&R routes is quantified to showcase the continuous repeatability of VirT\&R. 

\subsection{Experimental Platforms}
A Clearpath Warthog UGV \cite{Warthog}, shown in Fig. 1, is equipped with an Ouster OS1-128 LiDAR \cite{ouster_128} and a NovAtel SMART6 RTK-GPS system \cite{novatelGPS}. The Warthog is used to run LT\&R on the teach maps generated from NeRF results. The DJI Phantom 4 Pro \cite{DJIPhantom4} is used for aerial imaging and uses the standard 20 MP CMOS sensor camera with an 84\textdegree{} field of view. Images for NeRF training were taken at 2.5 Hz (approximately every 0.5 m), and relative transformations from virtual driving on the mesh were recorded at 20 Hz.

\begin{table}[b!]
    \renewcommand{\arraystretch}{1.25}
    \centering
    \caption{Experimental routes around UTIAS used to evaluate VirT\&R.}
    \begin{tabular}{|>{\centering\arraybackslash}p{1.8cm}|>{\centering\arraybackslash}m{2.5cm}|>{\centering\arraybackslash}m{2.5cm}|}
      \hline
      \rule{0pt}{13pt}\textbf{Path} & \textbf{VirT\&R Teach Distance} &  \textbf{\# and Distance VirT\&R Repeats} \\[6pt] \hline \hline
        UTIAS Parking & 350 m & 10 (3.5 km) \\
        \hline
        Mars Dome & 290 m & 10 (2.9 km) \\
        \hline
        UTP Parking  & 366 m & 10 (3.7 km) \\
        \hline
        UTP Survey & 463 m & 5 (2.3 km) \\
        \hline \hline
        \textbf{Total} & 1469 m & 35 (12.4 km) \\
        \hline
    \end{tabular}
    \label{tab:route_info}
    \vspace{-4mm}
\end{table}

\begin{figure}[b!]
\vspace*{1mm}
    \centering
    \includegraphics[width=1.0\columnwidth]{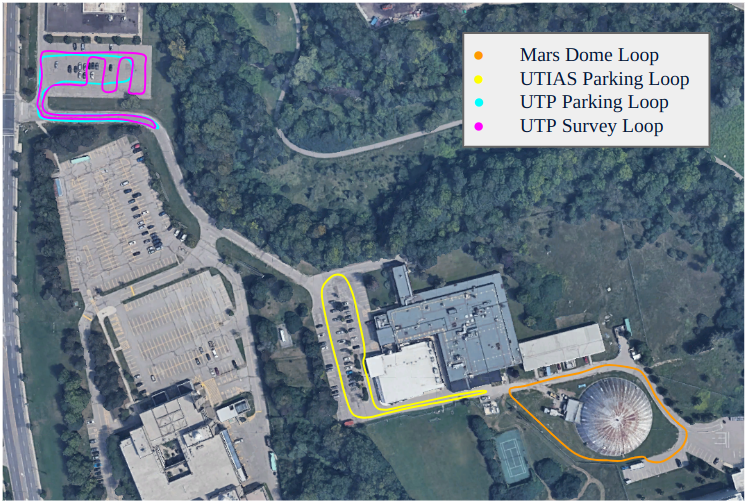}
    \caption{Visualizations of all four loops VirT\&R was evaluated on, consisting of urban settings with buildings, cars, and paved paths, as well as more off-road settings with vegetation, trees, fencing, and grassy paths.}
    \label{fig:loops}
\end{figure}

\subsection{Evaluation Routes} 
VirT\&R was tested on the four routes in Fig. \ref{fig:loops} at the University of Toronto Institute for Aerospace Studies (UTIAS) campus. The UTIAS Parking Loop and UTP Loops consist of entirely paved paths, buildings, and cars. The Mars Dome route consists of a partially grass/snow-covered path, slight elevation changes, more vegetation, and fencing. Detailed visual cues embedded in the NeRF models of these environments include salt stains, tire tracks, puddles, and shadows, which enable high-fidelity terrain assessment to be done by the pilot in simulation.

The Mars Dome was a particularly difficult environment due to 0.5 m of snow build-up, which required a path to be shoveled for the Warthog to be able to traverse part of the environment. This resulted in some parts of the route having less than 30 cm of lateral clearance during tight turns. Additionally, in the UTP Loop, two routes were taught to showcase one of the benefits of the VirT\&R: teaching multiple routes through one environment without any new scene data capture. Table \ref{tab:route_info} presents the environments and corresponding distances driven. 

A total of 12.4 km of autonomous driving was carried out across the four routes to demonstrate the consistent efficacy of VirT\&R, and qualitatively assess its performance. The Mars Dome and UTIAS Parking Loops are evaluated using spray paint markings to obtain an absolute lateral path-tracking error, as permission was obtained to mark the pavement only in those locations. Two example repeats for each of the four routes are also compared to each other to quantify T\&R's relative repeatability when using a virtual teach map.

\section{Results}
\label{sec:results}
Qualitative assessment of repeat performance on all routes reveals that the robot can reliably follow the virtual path over long distances, navigate narrow openings, complete tight turns, and consistently hit specific marks. The NeRF meshes provide sufficient detail to infer scene characteristics—including potential dynamic elements—and the point clouds accurately represent the environment, even in predominantly flat areas. The UGV follows a route smoothly and is resilient to both stationary and moving cars, pedestrians, and melted snow banks while repeating with a virtual teach map.

\subsection{LiDAR Teach and Repeat Baseline Comparison} 

To quantitatively compare VirT\&R to LT\&R, we implemented our physical marking-based evaluation methodology to obtain the lateral path-tracking RMSE. Measurements for the LiDAR and VirT\&R repeats were taken five times each for the Mars Dome and UTIAS Parking Lot Loops, using identical controller parameters. Due to the nature of physically measuring these marks, an uncertainty of $\pm$4 cm has been associated with the readings to account for the measuring tape’s precision as well as human error when placing the jig and stopping the Warthog next to the mark. Thus, we expect the LT\&R repeats to agree with recently published \cite{qiao_radar_2024} data within this uncertainty envelope.

\begin{table*}[t] 
\vspace*{1mm}

\renewcommand{\arraystretch}{1.25}
\caption{Estimated and measured RMSE and maximum path-tracking errors for VirT\&R and LT\&R across two different environments.}
\centering

\resizebox{\textwidth}{!}{ 
\begin{tabular}{|c|c|>{\centering\arraybackslash}m{2cm}|>{\centering\arraybackslash}m{2cm}|>{\centering\arraybackslash}m{2cm}|>{\centering\arraybackslash}m{2cm}|}
    \hline  
    \textbf{Modality} & \textbf{Path} & \textbf{T\&R-Estimated Lateral RMSE (m)} & \textbf{T\&R-Estimated Max Lateral Error (m)} & \textbf{Measured Lateral RMSE (m)} & \textbf{Measured Max Lateral Error (m)}  \tabularnewline \hline \hline
    \multirow[c]{2}{*}{LT\&R}  
    & UTIAS Parking  & 0.076 & 0.207 & 0.089 & 0.241 \tabularnewline \cline{2-6}
    & Mars Dome     & 0.056 & 0.267 & 0.109 & 0.279 \tabularnewline \hline
    \multirow[c]{2}{*}{VirT\&R (Ours)}  
    & UTIAS Parking  & 0.152 & 0.434 & 0.184 & 0.476 \tabularnewline \cline{2-6}
    & Mars Dome     & 0.189 & 0.581 & 0.195  & 0.394 \tabularnewline \hline
\end{tabular}
}
\label{tab:primaryResults}
\end{table*}

\begin{figure}[t]
    \centering
    \includegraphics[width=1.0\columnwidth]{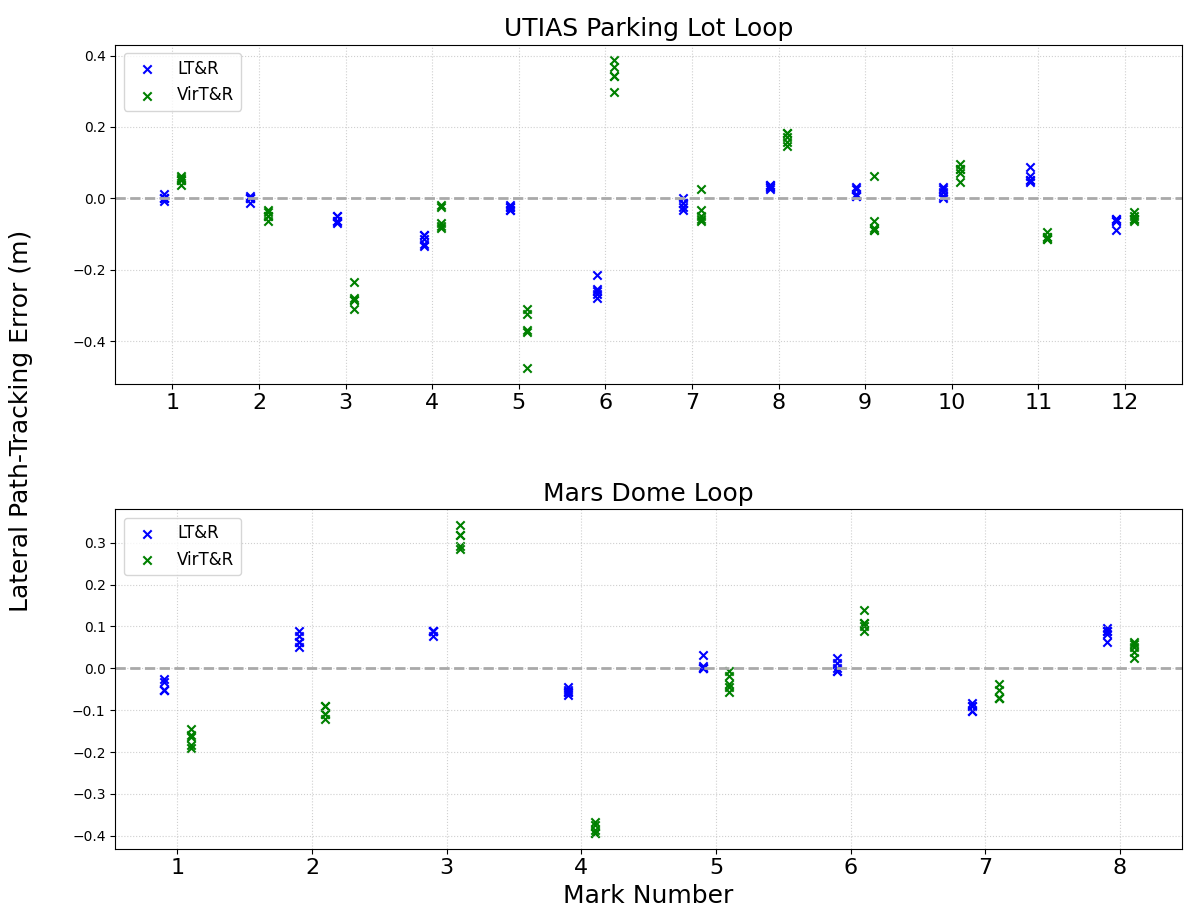} 
    \caption{Lateral path-tracking error distribution for measurements of each spray paint mark for the UTIAS Parking Lot and Mars Dome Loops taken over five repeats of both VirT\&R and LT\&R.}
    \label{fig:absolute}
\end{figure}

The marking offset measurements taken during VirT\&R and LT\&R repeats for the Mars Dome and UTIAS Parking Loops can be seen in Fig. \ref{fig:absolute}, where the Mars Dome Loop had eight marks, and the UTIAS Parking Loop had twelve. Although the offset is consistently greater for VirT\&R, the spread of the measurements is less than 10 cm for all marks, suggesting that VirT\&R's localization performs similarly to LT\&R, allowing for the UGV to repeat in its tracks—a finding we observed qualitatively throughout testing, as faint tracks were visible after going through snow and puddles.

A summary of the lateral RMSE and maximum error values for VirT\&R and LT\&R, determined through physical measurement, can be seen in Table \ref{tab:primaryResults}, which also lists the T\&R-estimated lateral and maximum errors obtained from the signed distances between every vertex in the teach and repeat pose graphs. The difference between the LT\&R estimated and measured values is within the uncertainty associated with this method and indicates that physically measuring lateral path-tracking error is a reliable assessment. Additionally, LT\&R estimated performance also aligns with a previous benchmark \cite{qiao_radar_2024}. The measured lateral RMSE for VirT\&R was 18.4 cm in the UTIAS Parking Loop and 19.5 cm in the Mars Dome Loop, which both fall within the range of LT\&R maximum errors. The estimated VirT\&R error for the Mars Dome Loop is likely higher than the measured value because the internal T\&R estimate considers all vertices in the pose graph of a repeat, and the spray paint mark was likely not at the specific vertex with the highest offset, particularly as it was intentionally driven close to. The maximum VirT\&R measured lateral path-tracking errors are 47.6 and 39.4 cm for the UTIAS Parking Loop and Mars Dome Loop respectively, signifying there is less accuracy with VirT\&R than with LT\&R, as expected. Overall, these results highlight that localization via NeRF-generated point clouds is a comparable method. 

\subsection{VirT\&R Repeatability}
The repeatability of VirT\&R is assessed by comparing two repeats each for the Mars Dome, UTIAS Parking Lot, and UTP Loops. This comparison is useful to gauge the affect that non-stationary objects and environmental changes have over time on the executed repeat. In lieu of GPS data, signed distances between vertices in the pose graph chains of respective repeats were used to determine the relative deviation between repeats to virtual teach maps. Given this same internally estimated metric was within the uncertainty of the measured lateral RMSE for the Mars Dome and UTIAS Parking Lot Loops, it is deemed to be a useful assessment.

\begin{figure}[t]
    \centering
    \includegraphics[width=1.0\columnwidth]{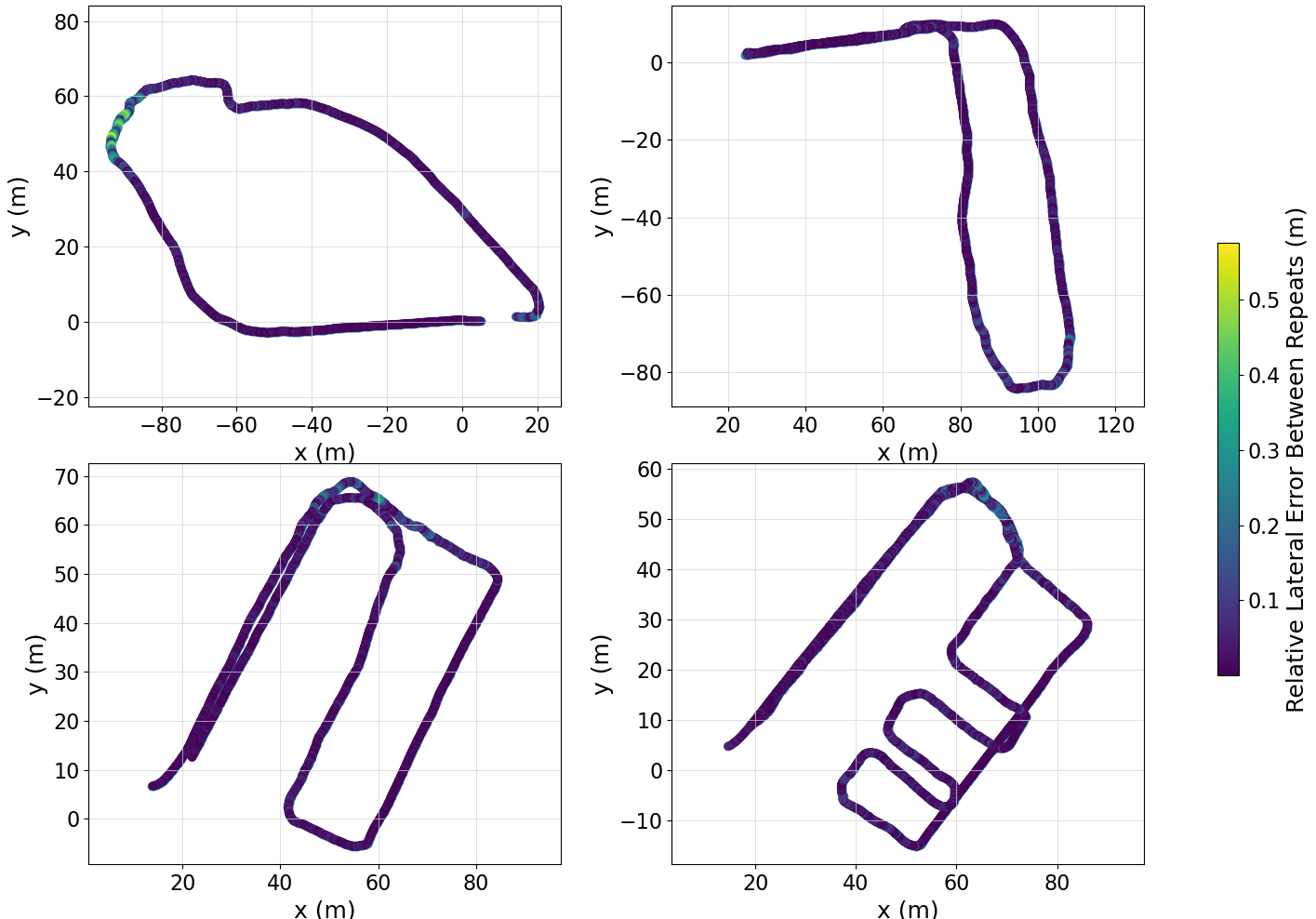} 
    \caption{VirT\&R repeatability is assessed by comparing two repeats made using virtual teach maps for each of the four routes. The relative path-tracking error between repeats is highlighted using colors ranging from purple (small error) to yellow (large error). Lateral error is observed to increase during tight turns or quick, precise maneuvering.}
    \label{fig:relative2}
\end{figure}

Across the four routes in Fig. \ref{fig:relative2}, the lateral error between two repeats increased from nearly zero to 55.4 cm, with the UTIAS Parking Loop exhibiting the best repeatability and the Mars Dome Loop having the worst. The portion of the Mars Dome Loop where the error is higher is also the area of the path with trees and slightly less structure. Additionally, during the virtual teach, intentionally rapid heading changes were driven here as a performance test. 

The maximum relative error for all routes is found in this swerving section of the Mars Dome Loop, near marking \#4, which has the highest measured absolute lateral path-tracking error for this loop of 39.4 cm, seen in Table \ref{tab:primaryResults}. The largest errors consistently occur along sharp or more frequent turns, suggesting that the NeRF-generated localization layer may not currently be robust enough to handle these maneuvers reliably. Overall, it can be seen that the lateral error between VirT\&R repeats is mostly small along the path and increases during tight turns or quick maneuvers.

\section{Discussion and Lessons Learned}
\label{sec:discussion}
Enabling autonomous control via the T\&R framework using virtually-generated submaps proved challenging in multiple respects. A significant observation was the difficulty in scene reconstruction with minimal noise and accurate geometry in challenging snowy conditions. NeRF model training noticeably suffered in environments where large swaths of untouched snow resulted in minimal detail and reduced depth information. Earlier aerial surveys of the same scenes produced much more detailed NeRF models when snow was not present. Interestingly, although VirT\&R was capable of running without issue in all the environments considered in this paper, attempts to use VirT\&R on the Grassy Loop—another route commonly used for T\&R testing \cite{qiao_radar_2024}—resulted in unpredictable localization in the flat, highly unstructured environment. This highlights a potential pitfall of the method and will be investigated further through the substitution of NeRF with other scene reconstruction techniques, such as classic photogrammetry. Additionally, the methodology used to evaluate the performance of VirT\&R is subject to inherent errors, both in how the UGV is paused to take measurements and in the positioning of the jig to measure the offsets. It is likely that if these errors could be reduced, the evaluated performance would more closely approach that of LT\&R. There remains room for improvement in the VirT\&R evaluation methodology, as the current experimental structure does not lend itself to being evaluated conventionally with GPS. To reduce the compounded errors in the `pseudo-GPS', fiducial tags, existing 3D models like Google Earth, or ground control points with known absolute GPS information could be used to further anchor the reconstruction in the same frame as GPS data taken during live VirT\&R repeats.

\section{Conclusion}
\label{sec:conclusion}
In this work, we presented Virtual Teach and Repeat: an autonomous route following stack capable of driving in various environments using only monocular commercial drone imagery and a piloted path in simulation as the only inputs to the system. The NeRFs of captured scenes allowed for rich point clouds and visually accurate meshes to be extracted and used for localization and pilot simulation. 
This extension to Teach and Repeat offers the advantages of constructing a large environment once—without requiring the physical presence of a human or UGV—and subsequently piloting any desired route within that environment without needing to capture it again. We compared the path-tracking performance of NeRF-derived submaps to LiDAR on four different routes and completed over 12 km of uninterrupted autonomous driving. VirT\&R performed well in the tested environments, achieving measured RMSE values of 19.5 cm and 18.4 cm and maximum errors of 39.4 cm and 47.6 cm. Relative repeat path analysis corroborates the qualitative observation that VirT\&R repeats its path smoothly, despite moving cars, pedestrians, and slight environmental changes. Finally, we intend to improve the methodology for assessing VirT\&R's accuracy and to test in increasingly unstructured environments. This will further enhance Virtual Teach and Repeat, expand cross-modal sensor support, and extend the capabilities of the Teach and Repeat framework.

\section{Acknowledgements}
This work was supported by the Natural Sciences and Engineering Research Council (NSERC). We thank Hexagon NovAtel and Clearpath Robotics for their support.




%
%
%


\bibliographystyle{./IEEEtranBST/IEEEtran}
\bibliography{./IEEEtranBST/IEEEabrv,./biblio}

\end{document}